\documentclass[10pt,twocolumn,letterpaper]{article}

\usepackage{cvpr}
\usepackage{times}
\usepackage{epsfig}
\usepackage{graphicx}
\usepackage{amsmath}
\usepackage{amssymb}

\usepackage{booktabs}
\usepackage{bm}
\usepackage{lipsum}
\usepackage{multirow}

\newcommand{\T}{^{\top}}    
\newcommand{\B}[1]{\textbf{#1}}    

\usepackage[pagebackref=true,breaklinks=true,colorlinks,bookmarks=false]{hyperref}

\cvprfinalcopy 

\def\cvprPaperID{1162} 

\ifcvprfinal\fi
\begin{document}

\title{Low-Shot Learning with Imprinted Weights}

\author{Hang Qi\thanks{The majority of the work was done while interning at Google.}\\
UCLA
\and
Matthew Brown\\
Google
\and
David G. Lowe\\
Google
}

\maketitle

\begin{abstract}
Human vision is able to immediately recognize novel visual categories after seeing just one or a few training examples. We describe how to add a similar capability to \mbox{ConvNet} classifiers by directly setting the final layer weights from novel training examples during low-shot learning. We call this process \emph{weight imprinting} as it directly sets weights for a new category based on an appropriately scaled copy of the embedding layer activations for that training example. The imprinting process provides a valuable complement to training with stochastic gradient descent, as it provides immediate good classification performance and an initialization for any further fine-tuning in the future. We show how this imprinting process is related to proxy-based embeddings. However, it differs in that only a single imprinted weight vector is learned for each novel category, rather than relying on a nearest-neighbor distance to training instances as typically used with embedding methods. Our experiments show that using averaging of imprinted weights provides better generalization than using nearest-neighbor instance embeddings. 

\end{abstract} 

\section{Introduction}

Human vision can immediately recognize new categories after a person is shown just one or a few examples~\cite{miller2000learning,lake2015human}. For instance, humans can recognize a new face from a photo of an unknown person and new objects or fine-grained categories from a few examples by implicitly drawing connections from previously acquired knowledge. Although deep neural networks trained on millions of images have in some cases exceeded human performance in large-scale image recognition~\cite{russakovsky2015imagenet}, under an open-world setting with emerging new categories it remains a challenging problem how to continuously expand the capability of an intelligent agent from limited new samples, also known as \textit{low-shot learning}.

Embedding methods~\cite{weinberger2009distance} have a natural representation for low-shot learning, as new categories can be added simply by pushing data examples through the network and performing a nearest neighbor algorithm on the result~\cite{schroff2015facenet}. It has long been realized in the semantic embedding literature that the activations of the penultimate layer of a \mbox{ConvNet} classifier can also be thought of as an embedding vector, which is a connection we further develop in this paper. \mbox{ConvNets} are the preferred solution for achieving the highest classification performance, and the softmax cross-entropy loss is faster to train than the objectives typically used in embedding methods, such as triplet loss.

In this paper, we attempt to combine the best properties of ConvNet classifiers\footnote{In this paper we use the term ``ConvNet classifiers'' to refer to convolutional neural networks trained with the softmax cross-entropy loss for classification tasks.} with embedding approaches for solving the low-shot learning problem. Inspired by the use of embeddings as proxies~\cite{movshovitz2017no} or agents~\cite{wang2017normface} for individual object classes, we argue that embedding vectors can be effectively compared to weights in the last linear layer of ConvNet classifiers. Our approach, called \textit{imprinting}, is to compute these activations from a training image for a new object category and use an appropriately scaled version of these activation values as the final layer weights for the new category while leaving the weights of existing categories unchanged. This is extended to multiple training examples by incrementally averaging the activation vectors computed from the new training images, which our experiments find to outperform nearest-neighbor classification as used with embedding approaches.

We consider a low-shot learning scenario where a learner initially trained on \textit{base classes} with abundant samples is then exposed to previously unseen \textit{novel classes} with a limited amount of training data for each category~\cite{hariharan2016low}. The goal is to have a learner that performs well on the combined set of classes. This setup aligns with human recognition which continuously learns new concepts during a lifetime.

Existing approaches exhibit characteristics that render them infeasible for resource-limited environments such as mobile devices and robots. For example, training a deep ConvNet classifier with stochastic gradient descent requires an extensive fine-tuning process that cycles through all prior training data together with examples from additional categories~\cite{hariharan2016low}. Alternatively, semantic embedding methods such as~\cite{oh2016deep,schroff2015facenet,sohn2016improved,song2017learning} can immediately remember new examples and use them for recognition without retraining. However, semantic embeddings are difficult to train due to the computationally expensive hard-negative mining step and these methods require storing all the embedding vectors of encountered examples at test time for nearest neighbor retrieval or classification. 

We demonstrate that the imprinted weights enable instant learning in low-shot object recognition with a single new example. Moreover, since the resulting model after imprinting remains in the same parametric form as ConvNets, fine-tuning via backpropagation can be applied when more training samples are available and when iterative optimization is affordable. Experiments show that the imprinted weights provide a better starting point than the usual random initialization for fine-tuning all network weights and result in better final classification results for low-shot categories. Our imprinting method provides a potential model for immediate recognition in biological vision as well as a useful approach for on-line updates for novel training data, as in a mobile device or robot.

The remainder of the paper is organized as follows. In Section~\ref{sec:related-work}, we discuss related work. Section~\ref{sec:connection} discusses the connections between embedding training and classification. Section~\ref{sec:imprinting} describes our approach. Then we provide implementation details and evaluate our approach with experiments in Sections~\ref{sec:implementation} and ~\ref{sec:experiments}. Section~\ref{sec:conclusion} concludes the paper.

\section{Related Work}  \label{sec:related-work}

\textbf{Metric Learning.}  Metric learning has been successfully used to recognize faces of new identities~\cite{chopra2005learning,schroff2015facenet} and fine-grained objects~\cite{movshovitz2017no,oh2016deep,rippel2015metric,song2016learnable,song2017learning}. The idea is to learn a mapping from inputs to vectors in an embedding space where the inputs of the same identity or category are closer than those of different identities or categories. Once the mapping is learned, at test time a nearest neighbors method can be used for retrieval and classification for new categories that are unseen during training. 

Contrastive loss~\cite{chopra2005learning} minimizes the distances between inputs with the same label while keeping the distances between inputs with different labels far apart. Rather than minimizing absolute distances, recent approaches formulate objectives focusing on relative distances. FaceNet~\cite{schroff2015facenet} optimizes a triplet loss and develops an online negative mining strategy to form triplets within a mini-batch. Instead of penalizing violating instance-based triplets independently, alternative loss functions regulate the global structure of the embedding space. Magnet loss~\cite{rippel2015metric} optimizes the distribution of different classes by clustering the examples using $k$-means and representing classes with centroids. Lifted structured loss~\cite{oh2016deep} incorporates all pair-wise relations within a mini-batch instead of forming triplets. The $N$-pair loss~\cite{sohn2016improved} requires each batch to have examples from $N$ categories for improved computational efficiency. All these methods require some online or offline batch generation step to form informative batches to speed up training. Structured clustering loss~\cite{song2016learnable} optimizes a clustering quality metric globally in the embeddings space.

The Proxy-NCA loss~\cite{movshovitz2017no} demonstrates faster convergence without requiring batch generation by assigning trainable proxies to each category, which we will describe in more detail in Section \ref{sec:connection}. NormFace~\cite{wang2017normface} explores a similar idea with all feature vectors normalized. The embedding can generalize to unseen categories, however the nearest neighbor model needs to store the embeddings of all reference points during testing. In our work, we retain the parametric form of ConvNet models and demonstrate that semantic embeddings can be used to imprint weights in the final layer. As a result, our approach has the same convergence advantages as ~\cite{movshovitz2017no} and~\cite{wang2017normface} during training, yet does not require storing embeddings for each training example or using nearest-neighbor search during inference.

\textbf{One-shot and Low-shot Learning.} One-shot or low-shot learning aims at training models with only one or a few training examples. The siamese network~\cite{koch2015siamese} uses two network streams to extract features from a pair of images and regress the inputs to a similarity score between two feature vectors. Matching networks~\cite{vinyals2016matching} learn a neural network that maps a small support set of images from unseen categories and an unlabeled example to its label. Prototypical networks~\cite{snell2017prototypical} use the mean embeddings of new examples as prototypes, but the embedding space is local with respect to the support classes due to the episodic scheme. These works formulate the low-shot learning problem as classifying an image among a number of unseen classes characterized by the support images; a query image and a support set must be provided together every time at inference. However, this evaluation setup does not align with human vision and many real-world applications where a learner grows its capability as it encounters more categories and training samples. In contrast, we consider an alternative setup similar to~\cite{hariharan2016low} which focuses on the overall performance of the learner on a combined set of categories including base classes represented by abundant examples together with novel low-shot classes. Hariharan and Girshick~\cite{hariharan2016low} train a multi-layer perceptron to generate additional feature vectors from a single example by drawing an analogy with seen examples. Their method retrains the last linear classifier at the low-shot training stage, whereas our approach allows instant performance gain on novel classes without retraining. More similar to our work is \cite{qiao2017few}, which trains parameter predictors for novel categories from activations. However, our method directly imprints weights from activations, which is made possible by architecture modifications that introduce a normalization layer.

\section{Metric Learning and Softmax Classifiers}  \label{sec:connection}

In this section, we discuss the connection between a proxy-based objective used in embedding training and softmax cross-entropy loss. Based on these observations, we then describe our method for extending ConvNet classifiers to new classes in the next section.

\subsection{Proxy-based Embedding Training}

Recent work has blurred the divide between triplet-based embedding training and softmax classification. For example, Neighborhood Components Analysis~\cite{goldberger2005neighbourhood} learns a distance metric with a softmax-like loss,
\begin{equation}
\mathcal{L}_{\textrm{NCA}}(x, y, Z) = - \log \frac{\exp(-d(x, y))}{\sum_{z\in Z} \exp(-d(x, z))}
\end{equation}
which makes points $x,y$ with the same label closer than examples $z$ with different labels under the squared Euclidean distance $d(x, y) = ||x - y||_2^2$. Movshovitz-Attias \etal~\cite{movshovitz2017no} reformulated the loss by assigning proxies $p(\cdot)$ to training examples according to the class labels
\begin{align}
\mathcal{L}_{\textrm{proxy}}(x)
\triangleq &\; \mathcal{L}_{\textrm{NCA}}(x, p(x), p(Z)) \nonumber\\
=& - \log \frac{\exp(-d(x, p(x)))}{\sum_{p(z)\in p(Z)} \exp(- d(x, p(z)))},
\label{eq:proxy-loss}
\end{align}
where $p(Z)$ is a set of all negative proxies. This formulation allows sampling anchor points $x$, rather than triplets, for each mini-batch and results in faster convergence than other objectives.

\subsection{Connections to Softmax Classifiers}

We will now discuss the connections between metric learning and softmax classifiers. We consider the case that each class has exactly one proxy and the proxy of a data point is determined statically according to its label. Concretely, let $C$ be the set of category labels and $P = \{p_1, p_2, \ldots, p_{|C|}\}$ be the set of trainable proxies, then the proxy of every point $x$ is $p(x) = p_{c(x)}$ where $c(x) \in C$ is the class label of $x$. 
We argue that the proxies $p_c$ are comparable to weights $w_c$ in softmax classifiers.

To see this, we assume point vectors and proxy vectors are normalized to the same length. It follows that minimizing the squared Euclidean distance between a point $x$ and its proxy $p(x)$ is equivalent to maximizing the inner-product, or equivalently cosine similarity, of the corresponding unit vectors
\begin{equation}
\min d(x, p(x)) \triangleq \min ||x - p(x)||_2^2 = \max x^{\top}p(x),
\end{equation}
since $||u - v||_2^2 = 2 - 2 u\T v$ for unit vectors $u, v \in \mathbb{R}^D$.
Substituting the squared Euclidean distance with inner product in Eq.~\ref{eq:proxy-loss}, the resulting loss can be written as
\begin{equation}
\mathcal{L}(x, c(x)) = - \log \frac{\exp(x^\top p_{c(x)})}{\sum_{c\in C} \exp(x^\top p_c)},
\end{equation}
which is comparable to the softmax cross-entropy loss used for training classifiers
\begin{equation}
\mathcal{L}_{\textrm{softmax}}(x, c(x)) = - \log \frac{\exp(x^\top w_{c(x)} + b_{c(x)})}{\sum_{c\in C} \exp(x^\top w_c + b_c)},
\end{equation}
with bias terms $b_c =0$ for all $c \in C$.

\section{Imprinting}  \label{sec:imprinting}

Given the conceptual similarity of normalized embedding vectors and final layer weights as discussed above, it seems natural that we should be able to set network weights for a novel class immediately from a single exemplar. In the following, we outline our proposed method to do this, which we call \emph{imprinting}. In essence, imprinting exploits the symmetry between normalized inputs and weights in a fully connected layer, copying the embedding activations for a novel exemplar into a new set of network weights.

To demonstrate this method, we focus on a two-stage low-shot classification problem where a learner is trained on a set of base classes with abundant training samples in the first stage and then grows its capability to additional novel classes for which only one or a few examples are available in the second stage.

\subsection{Model Architecture}

\begin{figure}[tbp]
\begin{center}
\includegraphics[width=\linewidth]{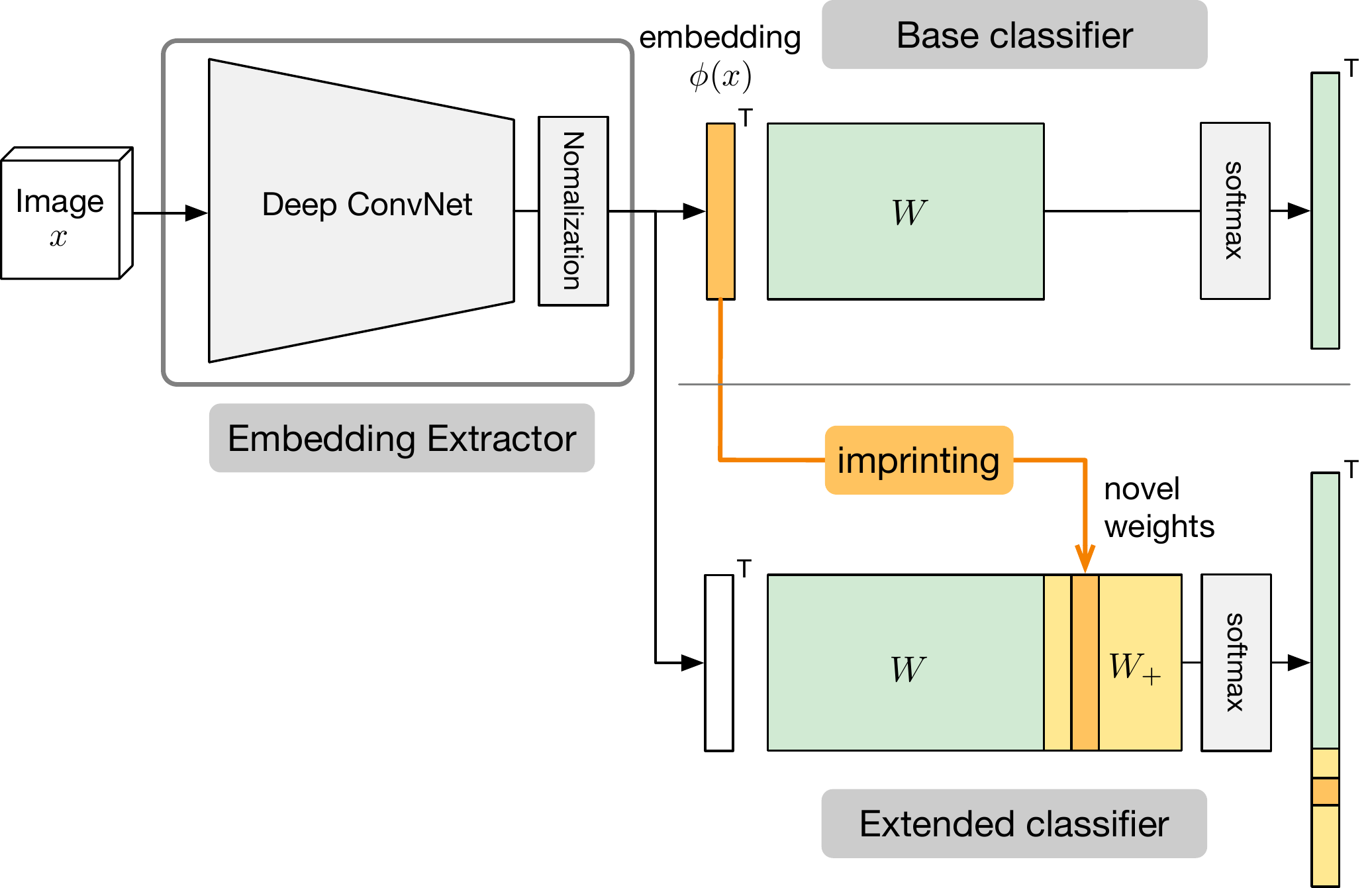}
\caption{The overall architecture of imprinting. After a base classifier is trained, the embedding vectors of new low-shot examples are used to imprint weights for new classes in the extended classifier.}
\label{fig:architecture}
\end{center}
\end{figure}

Our model consists of two parts. First, an embedding extractor $\phi: \mathbb{R}^{N} \rightarrow \mathbb{R}^{D}$, parameterized by a convolutional neural network, maps an input image $x \in \mathbb{R}^{N}$ to a $D$-dimensional embedding vector $\phi(x)$. Different from standard ConvNet classifier architectures, we add an $L_2$ normalization layer at the end of the embedding extractor so that the output embedding has unit length, \ie $||\phi(x)||_2 = 1$. Second, a softmax classifier $f(\phi(x))$ maps the embedding into unnormalized logit scores followed by a softmax activation that produces a probability distribution across all categories
\begin{equation}
f_i(\phi(x)) = \frac{\exp(w_i\T \phi(x))}{\sum_c \exp(w_c\T \phi(x))},
\label{eq:softmax}
\end{equation}
where $w_i$ is the $i$-th column of the weight matrix normalized to unit length. No bias term is used in this layer.

We view each column of the weight matrix as a template of the corresponding category. Unlike in~\cite{movshovitz2017no} where only the embedding extractor part is used during test time with the auxiliary proxies thrown away, we keep the entirety of the network. In the forward pass, the last layer in our model computes the inner product between the embedding of the input image $\phi(x)$ and all the templates $w_i$. With embeddings and templates normalized to unit lengths, the resulting prediction is equivalent to finding the nearest template in the embedding space in terms of squared Euclidean distance
\begin{equation}
\hat{y} = \arg\max_{c\in C} w_c\T \phi(x) = \arg\min_{c\in C} d(\phi(x), w_c).
\end{equation}
Compared with non-parametric nearest neighbor models, however, our classifier only contains one template per class rather than storing a large set of reference data points.

\textbf{Normalization.} Normalizing embeddings and columns of the weight matrix in the last layer to unit lengths is an important architectural design  in our model. Geometrically, normalized embeddings and weights lie on a high-dimensional sphere. In contrast, existing deep neural networks normally encourage activations to have zero mean and unit variance within mini-batches~\cite{ioffe2015batch} or layers~\cite{ba2016layer} for optimization reasons while they do not address the scale differences between neuron activations and weights. In our model, as a result of normalizing embeddings and columns of the weight matrix, the magnitude differences do not affect the prediction as long as the angle between the normalized vectors remains the same, since the inner product $w_i\T \phi(x) \in [-1, 1]$ now measures cosine similarity. Recent work in cosine normalization~\cite{luo2017cosine} discusses a similar idea of replacing the inner product with a cosine similarity for bounded activations and stable training, while we arrive at this design from a different direction. In particular, this establishes a symmetric relationship between normalized embeddings and weights, which enables us to treat them interchangeably.

\textbf{Scale factor.} The cosine similarity $w_i\T \phi(x) \in [-1, 1]$ can prevent the normalized probability of the correct class from reaching close to $1$ when applying softmax activation. For example, consider for an input $x$ the inner product producing $1$ for the correct category and producing the minimum possible value $-1$ for the incorrect categories, the normalized probability is $p(y_i | x) = e^{1} / [e^1 + (|C|-1)e^{-1}] = 0.069$, assuming a total of $|C| = 100$ categories. In consequence, it fails to produce a distribution close to the one-hot encoding of the ground truth label and therefore imposes a lower bound on the cross-entropy loss. This effect becomes more severe as the number of categories increases. To alleviate this problem, we adapt a scaling factor in our model as discussed by Wang~\etal~\cite{wang2017normface}. Concretely, we modify Eq.~\ref{eq:softmax} by adding a trainable scalar $s$ shared across all classes to scale the inner product
\begin{equation}
f_i(\phi(x)) = \frac{\exp(s w_i\T \phi(x))}{\sum_c \exp(s w_c\T \phi(x))}.
\end{equation}
We also experimented with the option of using an adaptive scale factor per class, but we did not observe significant effects on classification accuracy compared to our use of a single global scale factor.

In summary, our model architecture is similar to standard ConvNet classifiers except for two differences. The normalized embeddings and weights introduce a symmetric relationship that allows us to treat them interchangeably. The scaled inner product at the final layer enables training the entire model with the cross-entropy loss in the same way that standard ConvNet classifiers are trained. Next, we discuss how to extend such a classifier to novel categories by leveraging the symmetry between embeddings and weights.

\subsection{Weight Imprinting}

Inspired by the effectiveness of embeddings in retrieving and recognizing objects from unseen classes in metric learning, our proposed imprinting method is to directly set the final layer weights for new classes from the embeddings of training exemplars. Consider a single training sample $x_+$ from a novel class, our method computes the embedding $\phi(x_+)$ and uses it to set a new column in the weight matrix for the new class, \ie $w_+ = \phi(x_+)$. Figure~\ref{fig:architecture} illustrates this idea of extending the final layer weight matrix of a trained classifier by imprinting additional columns for new categories. 

\begin{figure}[htbp]
\begin{center}
\includegraphics[width=0.85\linewidth]{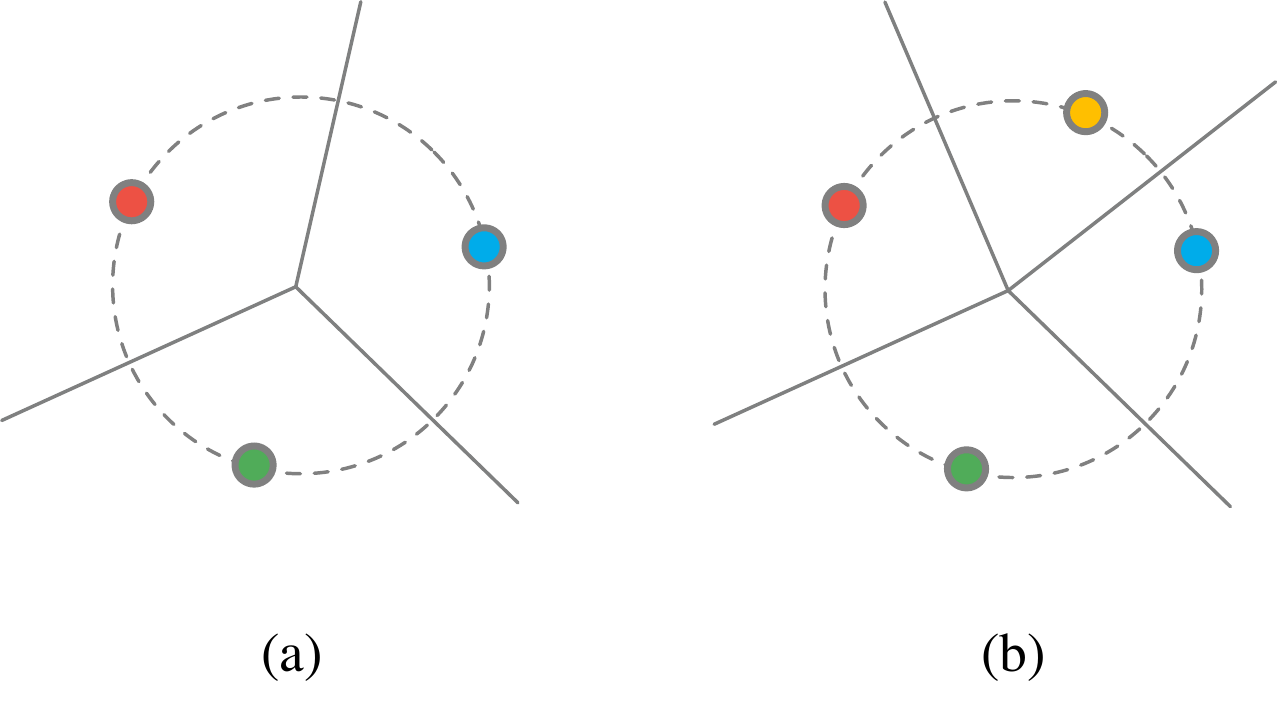}
\caption{Illustration of imprinting in the normalized embedding space. (a) Before imprinting, the decision boundaries are determined by the trained weights. (b) With imprinting, the embedding of an example (the yellow point) from a novel class defines a new region.}
\label{fig:embedding-boundary}
\end{center}
\end{figure}

Intuitively, one can think of the imprinting operation as remembering the semantic embeddings of low-shot examples as the templates for new classes. Figure~\ref{fig:embedding-boundary} illustrates the change of the decision boundaries after a new weight column is imprinted. The underlying assumption is that test examples from new classes are closer to the corresponding training examples, even if only one or a few are observed, than to instances of other classes in the embedding space. Notably, this desired property coincides with metric learning objectives such as triplet loss. The proxy-based loss, from which we have derived our method, upper bounds the instance-based triplet loss~\cite{movshovitz2017no}.

\textbf{Average embedding.} If $n>1$ examples $\{x_+^{(i)}\}_{i=1}^{n}$ are available for a new class, we compute new weights by averaging the normalized embeddings $\tilde{w}_+ = \frac{1}{n}\sum_{i=1}^{n}\phi(x_+^{(i)})$ and re-normalizing the resulting vector to unit length $w_+ = \tilde{w}_+ / ||\tilde{w}_+|| $. In practice, the averaging operation can also be applied to the embeddings computed from the randomly augmented versions of the original low-shot training examples.

\textbf{Fine-tuning.} Since our model architecture has the same differentiable form as ordinary ConvNet classifiers, a fine-tuning step can be applied after new weights are imprinted. The average embedding strategy assumes that examples from each novel class have a unimodal distribution in the embedding space. This may not hold for every novel class since the learned embedding space could be biased towards features that are salient and discriminative among base classes. However, fine-tuning (using backpropagation to further optimize the network weights) should move the embedding space towards having unimodal distribution for the new class.

\section{Implementation Details}  \label{sec:implementation}
The implementation details are comparable to~\cite{movshovitz2017no} and~\cite{oh2016deep}. For training, all the convolutional layers are initialized from ConvNet classifiers pre-trained on the \mbox{ImageNet} dataset~\cite{russakovsky2015imagenet}. InceptionV1~\cite{szegedy2015going} is used in our experiments. The parameters of the fully-connected layers producing the embedding and unnormalized logit scores are initialized randomly. $L_2$ normalization is used for embedding vectors and weights in the last layer along the embedding dimension. Input images are resized to 256$\times$256 and cropped to 224$ \times$224. Intensity is scaled to $[-1, 1]$. During training, we augment inputs with random cropping and random horizontal flipping. The learning rate is 0.0001 for pre-trained layers; a 10$\times$ multiplier is used for randomly initialized layers. We apply exponential decay every four epochs with decay rate 0.94. The RMSProp optimizer is used with momentum 0.9. During testing, input patches are cropped from the center.

\section{Experiments}  \label{sec:experiments}

We empirically evaluate the classifiers containing imprinted weights. We first describe the overall protocols, then we present results on the CUB-200-2011 dataset.

\subsection{Data Splits}

The CUB-200-2011 dataset~\cite{wah2011caltech} contains 200 fine-grained categories of birds with 11,788 images. We use the train/test split provided by the dataset. In addition, we treat the first 100 classes as base classes where all the training examples (about 30 images per class on average) from these categories are used to train a base classifier. The remaining 100 classes are treated as novel classes where only $n$ examples from the training split are used for low-shot learning. We experiment with a range of sizes $n = 1, 2, 5, 10, 20$ of novel exemplars for the low-shot training split. During testing, the original test split that includes both base and novel classes is used. We measure the top-1 classification accuracy of the final classifier on all categories. To show the effect of weight imprinting for low-shot categories, we also report the performance on the test examples from the novel classes only.

\subsection{Models and Configuration Variants}

\textbf{Imprinting.} To obtain imprinted models, we compute embeddings of novel examples and set novel weights in the final layer directly. When more than one novel example is available for a class, the mean of the normalized embeddings is used. The basic configuration (\textit{Imprinting}) uses only the novel examples in their original forms. Alternatively, we experiment with random augmentation (\textit{Imprinting+Aug}). Five augmented versions are generated for each novel example by random cropping and horizontal flipping, followed by averaging the embedded vectors. Both variants require only forward-pass computation of a trained embedding extractor without any iterative optimization. We compare these imprinting variants against a model initialization consisting of random novel weights without fine-tuning (\textit{Rand-noFT}), which also involves zero backpropagation. Random weights are generated with a Xavier uniform initializer~\cite{glorot2010understanding}.

\textbf{Fine-tuning.} To demonstrate that imprinted weights can be used as better initializations than random values, we apply fine-tuning to the imprinting model (\textit{Imprinting+FT}) and to the model with random novel weights (\textit{Rand+FT}), respectively. In both cases, we fine-tune the entire network end-to-end. We use only low-shot examples from novel classes in addition to all training examples from base classes. When the distribution across all classes is imbalanced, we oversample the novel classes such that all the classes are sampled uniformly for each mini-batch. Random data augmentation is also applied.

\textbf{Jointly-trained ConvNet classifier.} For comparison, we train a ConvNet classifier for base and novel classes jointly without a separate low-shot learning phase (\textit{AllClassJoint}). The same data splits and preprocessing pipeline are used as in the fine-tuning cases. This model does not normalize embeddings or weights.

\textbf{Other low-shot methods.} We also apply the feature generator~\cite{hariharan2016low} and matching networks~\cite{vinyals2016matching} to our normalized embeddings trained with the softmax loss for comparison.

\subsection{Results}

\begin{table}[tbp]
\begin{center}
\resizebox{\linewidth}{!}{
\begin{tabular}{clrrrrr}
\toprule
{} & \multicolumn{1}{r}{$n=$} &     1  &     2  &     5  &     10 &     20 \\
\midrule
\multirow{3}{*}{\rotatebox[origin=c]{90}{\parbox[c]{1cm}{\centering \small{w/o FT}}}}
& Rand-noFT\footnotemark    &  0.17  &  0.17   & 0.17   & 0.17   &  0.17  \\
& Imprinting        &  21.26 &  28.69  & \B{39.52}  & 45.77  &  49.32 \\ 
& Imprinting + Aug  &  \B{21.40} &  \B{30.03}  & 39.35  & \B{46.35}  &  \B{49.80} \\ 
\hline
\multirow{3}{*}{\rotatebox[origin=c]{90}{\parbox[c]{1cm}{\centering \small{w/ FT}}}}
& Rand + FT         &  5.25  &  13.41 &  34.95 &  54.33 &  65.60 \\
& Imprinting + FT   &  \B{18.67} &  \B{30.17} &  \B{46.08} &  \B{59.39} &  \B{68.77} \\
& AllClassJoint     &  3.89  &  10.82 &  33.00 &  50.24 &  64.88 \\  
\hline
& Generator + Classifier~\cite{hariharan2016low}  & 18.56 &  19.07 &  20.00 &  20.27 &  20.88 \\
& Matching Networks~\cite{vinyals2016matching}  &  13.45 &  14.75 &  16.65 &  18.81 &  25.77 \\
\bottomrule
\end{tabular}
}
\end{center}
\caption{200-way top-1 accuracy for novel-class examples in CUB-200-2011. Imprinting provides good immediate performance without fine tuning. Adding data augmentation (Imprinting+Aug) does not give significant further benefit. The second block of 3 rows shows the results of fine tuning, for which the imprinting initialization retains an advantage. This remains true even when compared to training all classes from scratch (AllClassJoint). The final 2 rows provide comparisons with previous methods.}
\label{tab:birds-eval-novel}
\end{table}

\footnotetext{Rand-noFT is listed for easy comparison. Strictly, the header $n=1,\ldots, 20$ does not apply, since low-shot examples are not used.}

\begin{table}[tbp]
\begin{center}
\resizebox{\linewidth}{!}{
\begin{tabular}{clrrrrr}
\toprule
{} & \multicolumn{1}{r}{$n=$} &     1  &     2  &     5  &     10 &     20 \\
\midrule
\multirow{3}{*}{\rotatebox[origin=c]{90}{\parbox[c]{1cm}{\centering \small{w/o FT}}}}
& Rand-noFT &  37.36 &  37.36 &  37.36 &  37.36 &  37.36 \\
& Imprinting        &  \B{44.75} &  48.21 &  \B{52.95} &  55.99 &  57.47 \\   %
& Imprinting + Aug  &  44.60 &  \B{48.48} &  52.78 &  \B{56.51} &  \B{57.84} \\    %
\hline
\multirow{3}{*}{\rotatebox[origin=c]{90}{\parbox[c]{1cm}{\centering \small{w/ FT}}}}
& Rand + FT         &  39.26 &  43.36 &  53.69 &  63.17 &  \B{68.75} \\
& Imprinting + FT   &  \B{45.81} &  \B{50.41} &  \B{59.15} &  \B{64.65} &  68.73 \\
& AllClassJoint     &  38.02 &  41.89 &  52.24 &  61.11 &  68.31 \\
\hline
 & Generator + Classifier~\cite{hariharan2016low}  &  45.42 &  46.56 &  47.79 &  47.88 &  48.22 \\
 & Matching Networks~\cite{vinyals2016matching}  & 41.71 &  43.15 &  44.46 &  45.65 &  48.63 \\
\bottomrule
\end{tabular}
}
\end{center}
\caption{200-way top-1 accuracy measured across examples in all classes (100 base plus 100 novel classes) of CUB-200-2011. Imprinting retains similar advantages for rapid learning and initialization of fine-tuning as seen in Table~\ref{tab:birds-eval-novel}.}
\label{tab:birds-eval-all}
\end{table}

\begin{figure}[tbp]
\begin{center}
\includegraphics[width=\linewidth]{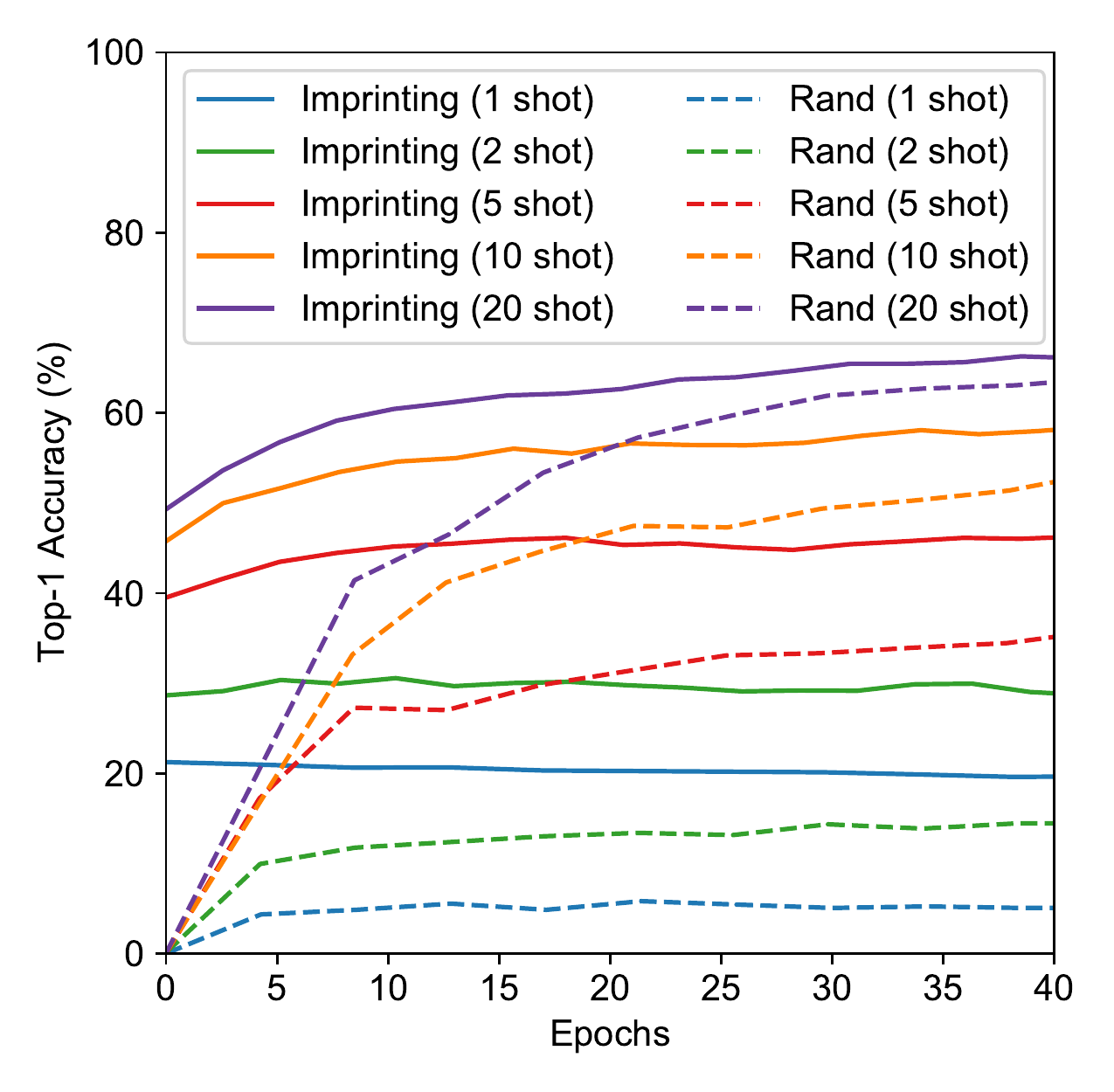}
\end{center}
\caption{Accuracy of fine-tuned models on novel classes for the first 40 epochs of training. Table~\ref{tab:birds-eval-novel} lists results after 112 epochs.}
\label{fig:finetune_novel}
\end{figure}

\begin{figure}[tbp]
\begin{center}
\includegraphics[width=\linewidth]{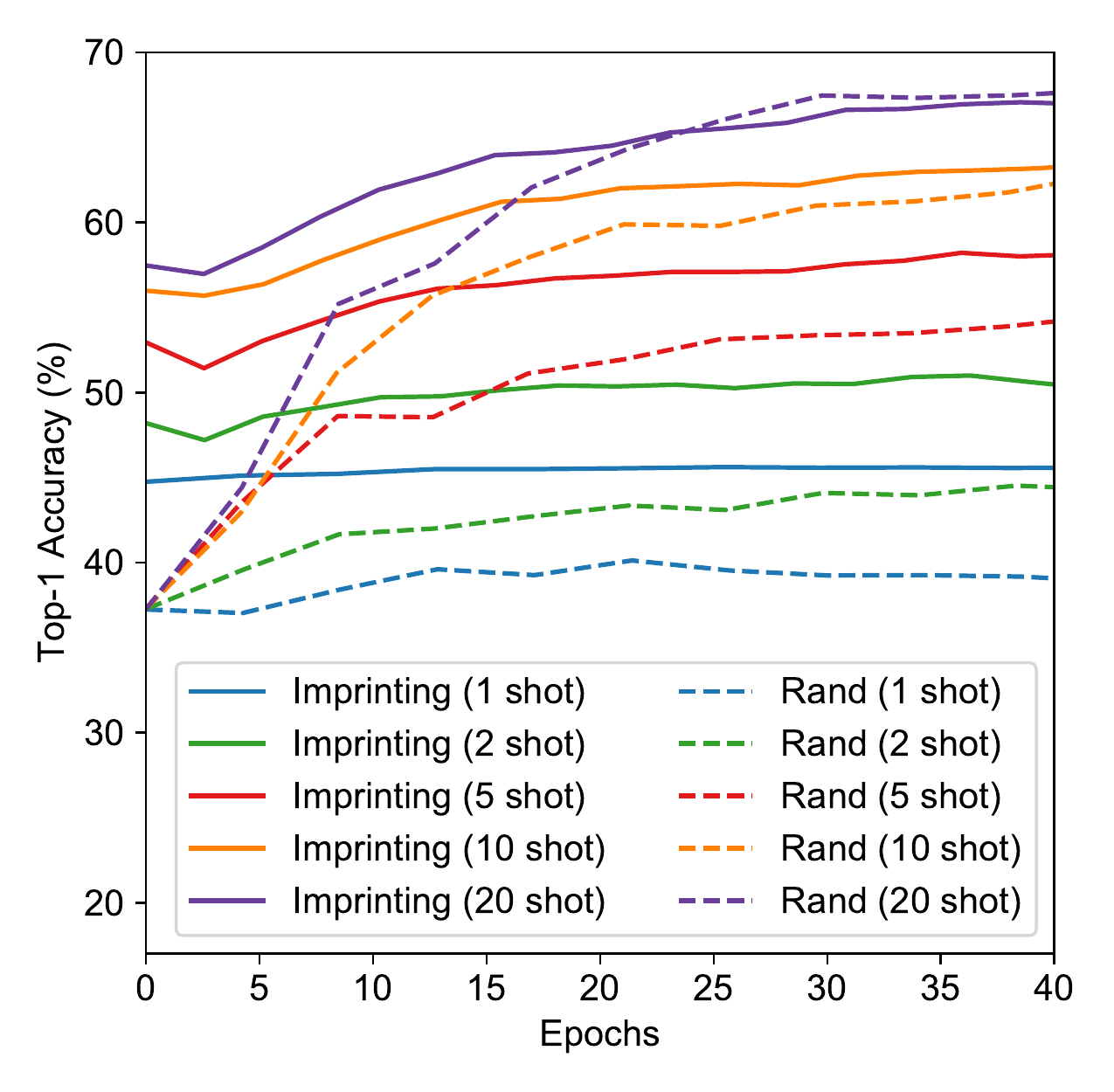}
\end{center}
\caption{Accuracy of fine-tuned models measured over all classes (100 base plus 100 novel classes) for the first 40 epochs of training. Table~\ref{tab:birds-eval-all} lists results after 112 epochs.}
\label{fig:finetune_all}
\end{figure}

\begin{figure*}[tbp]
\begin{center}
\includegraphics[width=\linewidth]{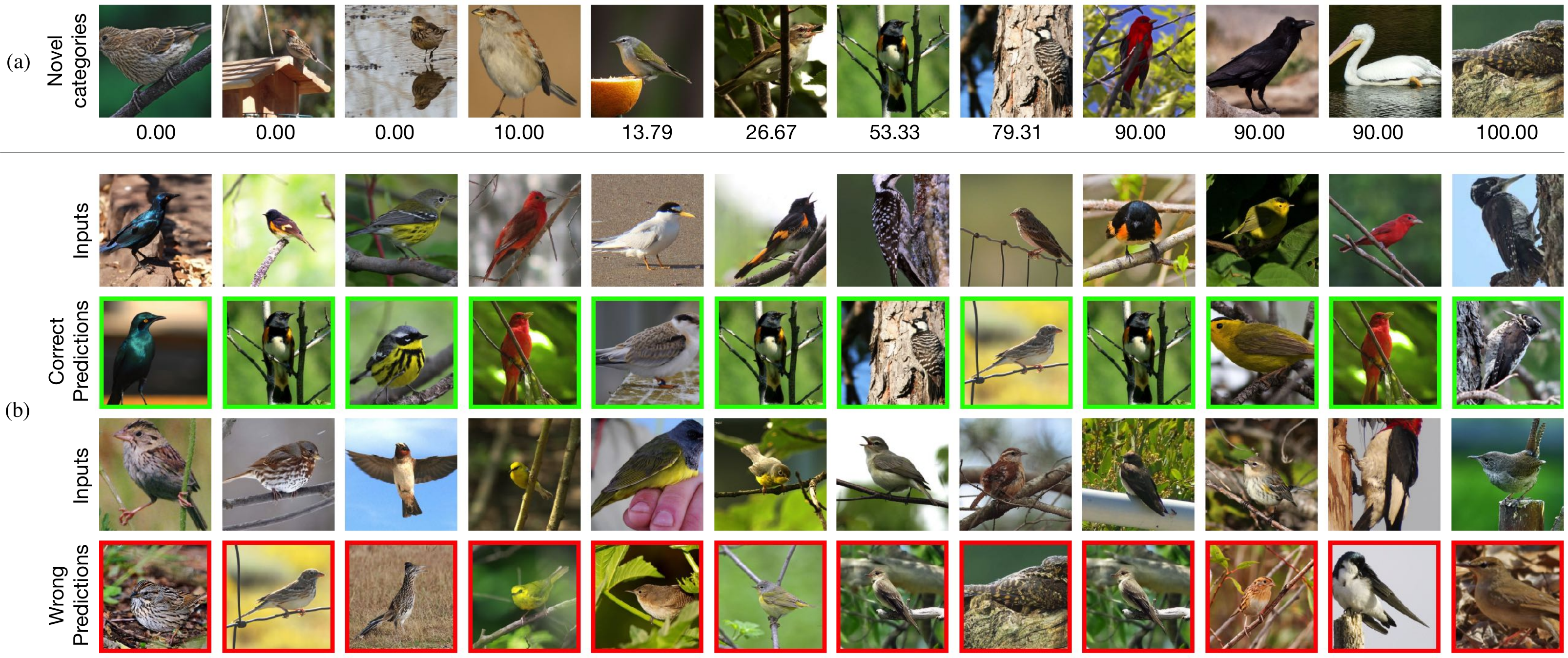}
\end{center}
\caption{(a) A subset of exemplars used for 1-shot training of novel classes sorted by their recall@1 scores as shown below each exemplar. High-performing categories tend to exhibit more distinctive colors, shapes, and/or textures. (b) Randomly selected success and failure cases predicted by a 1-shot imprinted model on CUB-200-2011. Test images and the 1-shot exemplar whose embedding was used to imprint the predicted class are shown in separate rows. Correct and wrong predictions are marked with green and red borders, respectively.}
\label{fig:birds-examples}
\end{figure*}

Tables~\ref{tab:birds-eval-novel} and~\ref{tab:birds-eval-all} show the top-1 accuracy of 200-way classification for novel examples and all examples in CUB-200-2011, respectively. Without any backpropagation, the imprinted weights computed from one-shot examples instantly provide good classification performance: 21.26\% on novel classes and 44.75\% on all classes. Imprinting using the average of multiple augmented exemplars (Imprinting+Aug), using the same random flips and crops as for base class training, does not give a significant improvement in performance. We conjecture this is because the embedding extractor has been trained on the base classes to be invariant to the applied transformations.

When fine-tuning the network weights with backpropagation, 
models initialized with imprinted weights (Imprinting+FT) take less time to converge and achieve better final accuracies than randomly initialized models, especially when limited low-shot examples are used. Figures~\ref{fig:finetune_novel} and~\ref{fig:finetune_all} plot evaluation accuracy of the fine-tunned models in the first 40 epochs on novel classes and all classes, respectively. Accuracies in Tables~\ref{tab:birds-eval-novel} and~\ref{tab:birds-eval-all} are recorded after around 112 epochs. For cases $n=1, 2$, the performance of imprinted weights is close to saturation and fine-tuning for more epochs can lead to degraded evaluation accuracies on novel classes, which we conjecture is due to overfitting on the 1 or 2 examples. The results show that the imprinted initialization can lead to better results for low-shot categories even when training from scratch on the entire dataset, as with AllClassJoint.

The classifier using generated features~\cite{hariharan2016low} has a similar performance to imprinting for $n=1$. While the matching network outperforms the feature generator as $n$ increases, we observe a performance gap when compared with imprinting. For our tests we modified the matching network to perform 200-way classification instead of 5-way~\cite{vinyals2016matching}.

Figure~\ref{fig:birds-examples} shows some sampled results following training of novel categories from the 1-shot imprinted model on CUB-200-2011. The top row shows randomly selected novel categories sorted by their classification accuracy as given below each exemplar. As might be expected, the highest-performing categories tend to exhibit more distinctive features of color, texture, and/or shape. In Figure~\ref{fig:birds-examples}(b) we show randomly selected success and failure cases predicted by the 1-shot imprinted model. The learned embeddings demonstrate an ability to generalize to different viewpoints, poses, and backgrounds from the single training example for each new category. 

\textbf{Transfer Learning with Imprinted Weights.} We show that imprinting benefits transfer learning in general. To transfer a trained classifier to a new set of classes, we substitute the final layer parameters with the mean embeddings of examples from new classes. The only difference between our approach and standard transfer learning approaches is that we initialize the new weights by imprinting rather than with random values. Note that the imprinting process requires little cost in terms of computation. Table~\ref{tab:birds-eval-novel-transfer} shows the top-1 classification accuracy of the imprinted model on the new classes. Random initialization yields an accuracy of 0.85\% while the models using imprinted weights have accuracies from 26.76\% up to 52.25\% as the number of training examples increases. Applying random augmentation (Imprinting+Aug) does not impact the performance significantly. Additional fine-tuning improves the performance. When novel training data is scarce ($n=1, 2, 5$), starting from the imprinted weights (Imprinting+FT) outperforms fine-tuning from random weights (Rand+FT) by a large margin. With more training examples, fine-tuning from imprinted weights converges to similar accuracy as when starting from random weights.

\begin{table}[tbp]
\begin{center}
\resizebox{\linewidth}{!}{
\begin{tabular}{clrrrrr}
\toprule
{} & \multicolumn{1}{r}{$n=$} &     1  &     2  &     5  &     10 &     20 \\
\midrule
\multirow{3}{*}{\rotatebox[origin=c]{90}{\parbox[c]{1cm}{\centering \small{w/o FT}}}}
& Rand-noFT         &  0.85  &  0.85   & 0.85   &  0.85   &  0.85  \\
& Imprinting        &  \B{26.76} &  33.11 &  43.00  &  48.74  &  52.25 \\
& Imprinting + Aug  &  26.08 &  \B{34.13} &  \B{43.34}  &  \B{48.91}  &  \B{52.94} \\
\hline
\multirow{2}{*}{\rotatebox[origin=c]{90}{\parbox[c]{0.8cm}{\centering \small{w/ FT}}}}
& Rand+FT           &  15.90 &  28.84 &  46.21 &  61.37 &  \B{71.57} \\ 
& Imprinting + FT   &  \B{26.59} &  \B{34.33} &  \B{49.39} &  \B{61.65} &  70.07 \\
\bottomrule
\end{tabular}
}
\end{center}
\caption{Top-1 accuracy for transfer learning on CUB-200-2011 using 1--20 examples for computing imprinted weights. The imprinted weights provide good immediate performance while also providing better final classification accuracy for 1 to 5 shot learning following fine tuning.}
\label{tab:birds-eval-novel-transfer}
\end{table}

\begin{figure}[tbp]
\begin{center}
\includegraphics[width=\linewidth]{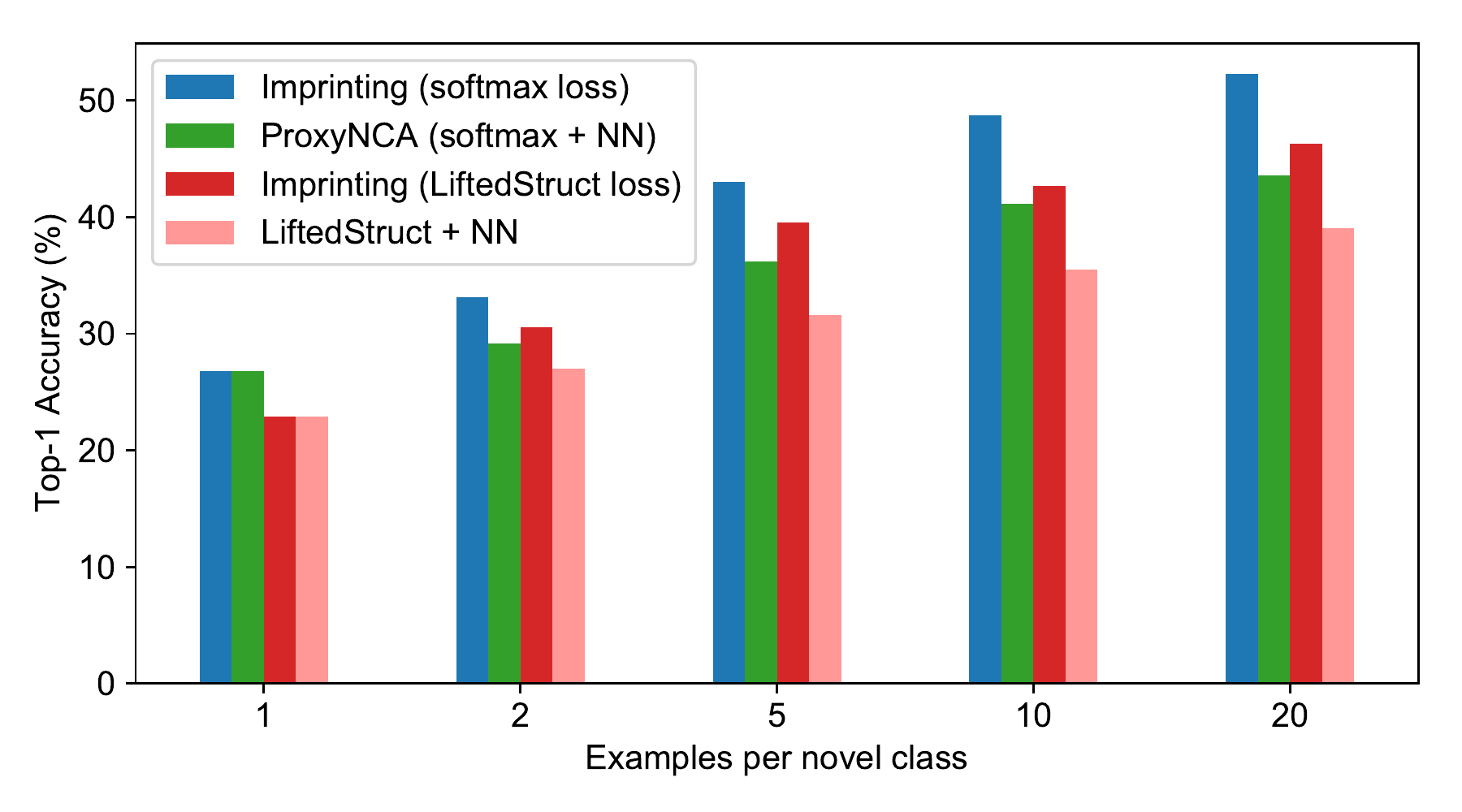}
\end{center}
\caption{Top-1 accuracy of 100-way classification on novel classes of CUB-200-2011. Imprinting averaged embeddings with a softmax loss (blue bars) outperforms storing all individual embeddings with a nearest-neighbor classifier (green). By comparison, embeddings trained with the lifted structured loss do not perform as well as with the softmax loss (red and pink). }
\label{fig:imprint-vs-knn}
\end{figure}

\textbf{Comparison with Nearest Neighbors.} 
As discussed in Section~\ref{sec:related-work}, the usual approach used in metric learning has been to store all exemplar embeddings and use the nearest neighbor algorithm for classification. Therefore, we compare our approach of using averaged embeddings with using a nearest-neighbor classifier where the embeddings of $n$ low-shot training examples from each novel class form the population set. When there is only one training example per class, $n=1$, the imprinted classifier is equivalent to the nearest neighbor classifier. When $n > 1$, the size of the imprinted classifier remains constant, whereas the size of the nearest neighbor classifier grows linearly as $n$ increases. Note that storing all embeddings trained with the softmax loss in a nearest-neighbor classifier is equivalent to a special case of Proxy-NCA~\cite{movshovitz2017no} using one proxy per class.

Perhaps surprisingly, the averaged embeddings perform better than storing all individual embeddings (Figure~\ref{fig:imprint-vs-knn}). We conjecture that the averaging operation reduces potentially noisy dimensions in the embedding to focus on those that are more consistent for that category. Although the averaging may not seem to be the optimal choice in cases where the distribution of novel class examples has multiple modalities in the embedding space, we do not observe this in our experiments. When the embedding space is first trained on the base classes, lower layers of the network will have been trained to bring multiple modalities together for feature inputs to the final linear layer. Moreover, keeping a single embedding for each class in the imprinted classifier has additional benefits since this standard form allows fine-tuning the embedding space with backpropagation and reduces test time computation and memory requirements.

\textbf{Comparison with Lifted Structured Loss.} Imprinted weights and Proxy-NCA are both trained with the softmax cross-entropy loss. Alternatively, we compare with embeddings trained with the lifted structured loss~\cite{oh2016deep}, which is a generalization of the widely used triplet loss. Figure~\ref{fig:imprint-vs-knn} shows that the softmax loss performs better in our experiments than the lifted structured loss. However, the lifted structured loss can also benefit from imprinting averaged embeddings rather than a nearest-neighbor classifier.

\begin{figure}[tbp]
\begin{center}
\includegraphics[width=\linewidth]{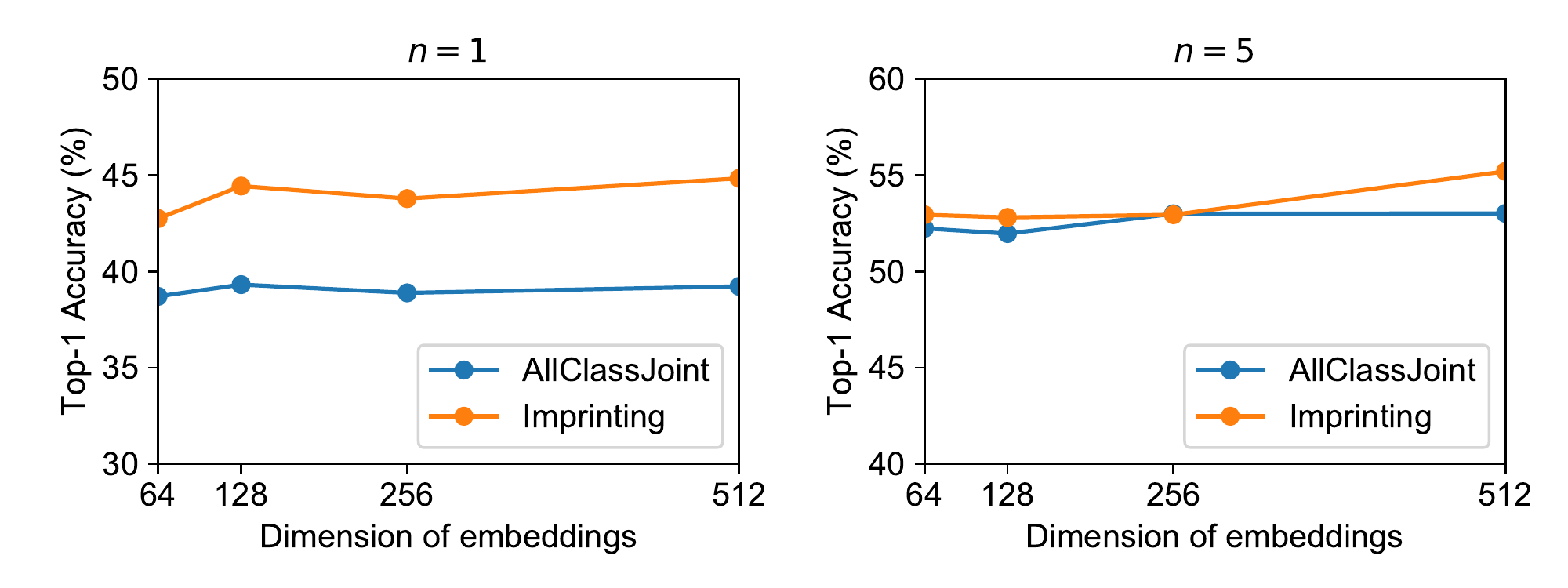}
\end{center}
\caption{Classification accuracies for Imprinting and AllClassJoint with different embedding dimensionalities under 1-shot and 5-shot settings, respectively.}
\label{fig:dimensionality}
\end{figure}

\textbf{Embedding Dimensionality.} We use 64-dimensional embeddings in all the experiments above. Empirically we experimented with various settings $D=64, 128, 256, 512$ for the imprinting model and the jointly-trained ConvNet Classifier (Figure~\ref{fig:dimensionality}). Increasing the dimensionality does not appear to have significant effects on the results. 

\section{Conclusions}  \label{sec:conclusion}

This paper has presented a new method, \emph{weight imprinting}, that directly sets the final layer weights of a \mbox{ConvNet} classifier for novel low-shot categories. This is a valuable complement to stochastic gradient descent, as it provides instant good classification performance on novel categories while allowing for further fine tuning when time permits. The key change that is made to the ConvNet architecture is a normalization layer with a scaling factor that allows activations computed for novel training examples to be directly copied (imprinted) as final layer weights. When multiple low-shot examples are presented, the computed activations for additional examples are averaged with the existing weights, which our experiments show to perform better than the nearest-neighbor approach typically used with embedding methods. An area for future research is whether the imprinting approach can also be used for more rapid training of other network layers, such as when encountering novel lower-level features.

{\small
\bibliographystyle{ieee}
\bibliography{imprint}
}

\end{document}